\documentclass[]{academic_template}

\usepackage[toc,page,header]{appendix}

\usepackage{amssymb}
\usepackage{amsmath}

\usepackage{makecell}
\usepackage{algorithm}      
\usepackage{algpseudocode}  
\usepackage{hyperref}
\usepackage{enumitem}
\usepackage[most]{tcolorbox}
\usepackage{xcolor}

\usepackage{graphicx}

\usepackage{listings}
\usepackage{url}
\usepackage{booktabs}
\usepackage{wrapfig}

\usepackage{colortbl}
\usepackage{xcolor}

\usepackage{caption}
\usepackage{subcaption}

\setlogoheight{14mm}   
\setlogospacing{5mm}
\setsjtublue 

\settitlerulethickness{3pt}  

\abstractboxon 
\setabstractframecolor{gray} 
\setabstractbgcolor{gray!10} 
\setlogotolineshift{5mm} 

\settoprulethickness{2.5pt}  
\setbottomrulethickness{1.5pt}

\title{DMRL: Document-Mediated Reinforcement Learning for Skill Optimization in Advertising Recommendation}

\author[1,*]{Wei Zhang}
\author[2,*]{Hongji Li}
\author[2]{Song Sun}
\author[2, \dagger]{Peng Yu}
\author[1, \dagger]{Xue Yang}
\author[2]{Lei Zhao}
\author[2]{Peng Jiang}

\affiliation[1]{Shanghai Jiao Tong University}
\affiliation[2]{Kuaishou Technology}

\contribution[*]{Equal contribution}
\contribution[\dagger]{Corresponding Author}

\abstract{
Advertising recommendation requires continuously tuning complex system parameters while balancing commercial returns and user experience. Recent work has introduced large language models (LLMs) with skill documents to assist this labor-intensive process, but skill optimization remains largely prompt-driven, lacking a principled mechanism to attribute rewards to specific document edits. To address this limitation, we propose Document-Mediated Reinforcement Learning (DMRL), a skill self-evolution framework that models skill document optimization as a sequence of structured editing actions. In DMRL, an upper-level agent performs controlled document edits, while a frozen lower-level task agent evaluates their effects through A/B testing. To address credit assignment and long-term outcomes, we introduce two key components: (1) Dual-Relative Policy Optimization (DRPO), a post-training policy optimization method for robust and risk-aware advantage estimation; and (2) Long-term Reward Predictor (LRP), which estimates long-term outcomes by modeling population heterogeneity with disentangled representation learning and cross-attention transfer. DMRL was deployed on a large-scale short-video ads platform and extensive empirical evaluation shows that DMRL outperforms state-of-the-art baselines across key advertising metrics.
}

\date{\today}

\begin{document}
\maketitle


\section{Introduction}
\label{sec:intro}


\begin{figure}[!tb]
  \centering
  \includegraphics[width=\linewidth]{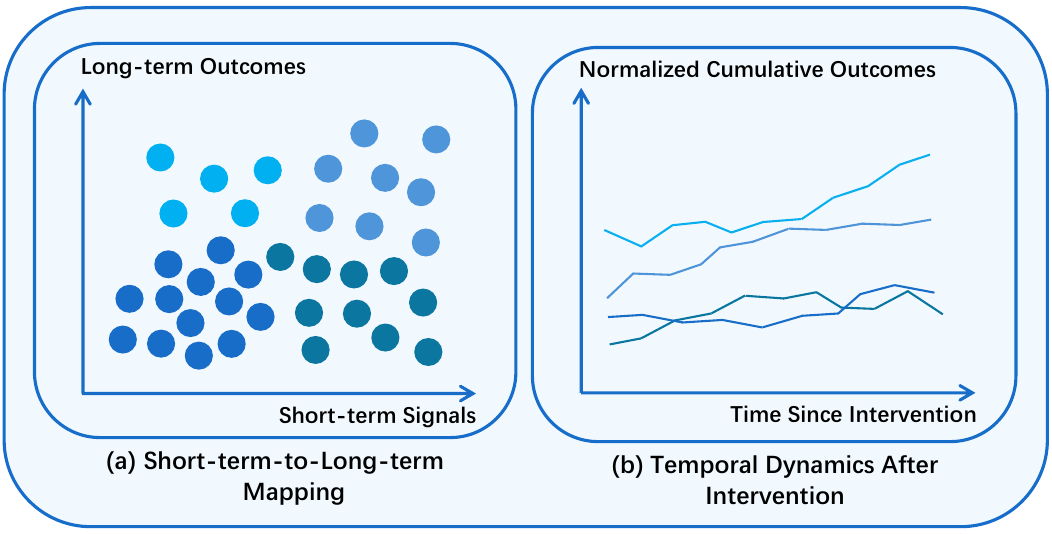}
  \caption{Substantial population heterogeneity in advertising. (a) Different user segments exhibit substantially different mappings from short-term signals to long-term outcomes. Short-term signals are measured within 1 day after intervention, while long-term outcomes are defined within 7 days. (b) Different user segments exhibit distinct temporal dynamics of normalized cumulative outcomes after intervention.}
  \label{fig:1_population}
\end{figure}

\begin{figure}[!tb]
  \centering
  \includegraphics[width=\linewidth]{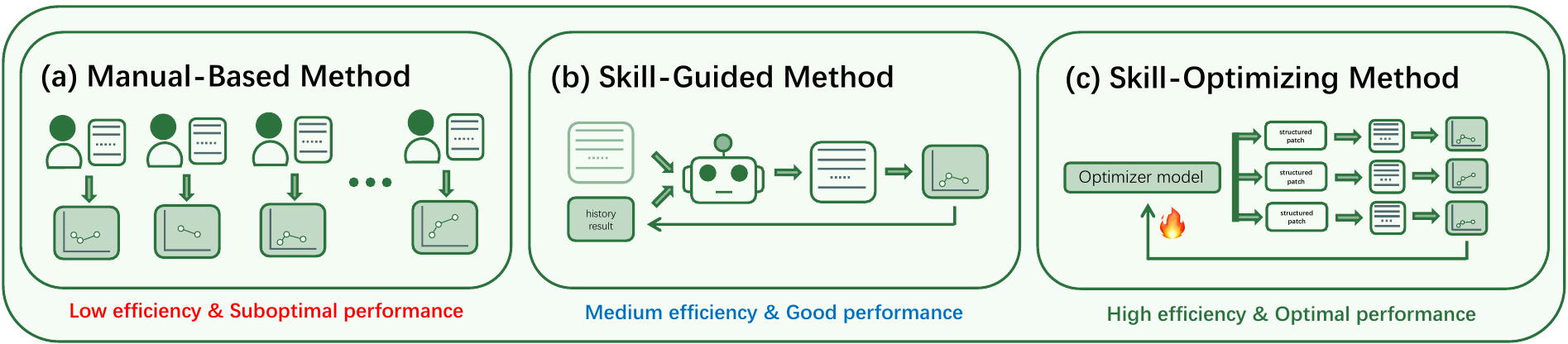}
  \caption{Comparison of the different parameter tuning paradigms in advertising recommendation system.}
  \label{fig:2_method}
\end{figure}

Online advertising recommendation depends on continuously balancing commercial returns and user experience \cite{yang2022click}. As illustrated in Figure~\ref{fig:2_method}(a), manual-based method requires substantial human effort and computational resources. Although hyperparameter optimization (HPO) provides a general framework \cite{bischl2023hyperparameter}, this process still relies heavily on domain experts with empirical intuition and iterative experimentation. Despite providing incremental gains, it suffers from three fundamental limitations. First, the human effort required scales linearly with the number of parameters, creating a significant operational bottleneck. Second, tuning experience is difficult to formalize and transfer across practitioners. Third, the trial-and-error process cannot guarantee systematic exploration of the parameter space, converging to locally optimal configurations.

The emergence of LLM-based agents offers a potential remedy to limitations above. Recent work has shown that large language models can assist with complex operational workflows by externalizing domain expertise into reusable skill documents~\cite{li2026skillsbench, jiang2026sok}, which specify task procedures, decision rules, and tool usage patterns. In advertising recommendation systems, such skill documents provide a promising interface for encoding tuning knowledge in a form continuously that is both executable by agents and reusable across scenarios. However, as illustrated in Figure~\ref{fig:2_method}(b), existing approaches still optimize such skills primarily through textual refinement or prompt-like updates~\cite{yang2026skillopt,alzubi2026evoskill,zhang2026coevoskills,qiu2026autorefine, wu2026agenticrectune}, without a principled mechanism for attributing observed rewards to specific document edits. We refer to this paradigm as skill-guided method.

Beyond edit-level credit assignment, a more fundamental challenge arises from the long-term outcomes. In advertising parameter tuning, the ultimate objective typically unfolds after days~\cite{chapelle2014modeling}, making it impractical to directly optimize the upper-level policy. Prior studies on long-term outcomes ~\cite{yu2024addressing,mcdonald2023impatient} offer partial solutions, but they do not explicitly address the pronounced population heterogeneity in advertising recommendation systems. As shown in Figure~\ref{fig:1_population}(a), the relationship between short-term signals and long-term outcomes differs substantially across user segments. Consequently, a predictor trained on aggregated data is dominated by majority populations, leading to biased estimation and poor generalization for underrepresented groups. Furthermore, Figure~\ref{fig:1_population}(b) reveals that outcomes trajectories after intervention also exhibit distinct temporal patterns across populations. Together, these observations suggest that long-term outcomes modeling should be population-aware rather than relying on a single predictor.

As illustrated in Figure~\ref{fig:2_method}(c), we propose DMRL, which moves beyond manual-based and skill-guided method by explicitly decomposing skill optimization into long-term outcomes estimation and edit-level reward attribution. We refer to this paradigm as skill-optimizing method. Specifically, DMRL's core consists of three synergistic modules: (1) a structural decoupling mechanism that separates an upper-level skill optimizer responsible for document modification from a lower-level task agent responsible for parameter intervention, with the skill document serving as the semantic interface between the two levels. (2) Dual-Relative Policy Optimization (DRPO), a post-training policy optimization method for robust and risk-aware advantage estimation that handles outcomes outliers, exploits treatment-control experimental structure, and regularizes edits with edit cost. (3) Long-term Reward Predictor (LRP), a prediction module that estimates long-term outcomes from short-term signals, using disentangled representation learning and cross-attention based historical transfer. These modules are optimized through a two-stage training strategy, designed to stabilize long-term outcomes prediction and ensure accurate advantage estimation. By combining structural decoupling of long-term outcomes prediction with robust advantage estimation, DMRL provides a principled and industrially viable solution for skill optimization on advertising recommendation platforms.

Our contributions are summarized as follows:

\begin{itemize}
    \item We propose DMRL, a novel skill optimization framework that connects skill document editing to downstream parameter optimization through structured semantic interfaces.
    \item We propose DRPO, a post-training policy optimization algorithm that extends GRPO by introducing a dual-relative advantage estimator and LRP, a long-term reward prediction module that mitigates high latency and population heterogeneity.
    \item We deploy the framework on a real-world advertising platform and demonstrate significant online improvements.
\end{itemize}

\section{Related Work}
\label{sec:related}


\subsection{Skill Acquisition and Optimization}
The idea of externalizing operational knowledge to guide agent behavior has emerged across several related lines of work. Early studies on tool use and action generation in LLM agents, such as ReAct~\cite{yao2022react} and Toolformer~\cite{schick2023toolformer}, show that exposing models to explicit interaction procedures or external APIs can substantially improve task performance. More recent work has further systematized such external knowledge as reusable agent skills~\cite{li2026skillsbench, jiang2026sok}. In terms of skill acquisition, Voyager~\cite{wang2023voyager} accumulates executable skills through lifelong interaction in an open-ended environment, while AgentTuning~\cite{zeng2024agenttuning} learns reusable agent behaviors from interaction data. Trace2Skill~\cite{ni2026trace2skill} instead distills structured skills directly from complete execution trajectories, converting passive episodic records into reusable procedures. A complementary line of work constructs skills from pre-existing resources rather than open-ended exploration: SkillFoundry~\cite{shen2026skillfoundry} and AutoSkill~\cite{yang2026autoskill} synthesize skill packages from domain knowledge bases to improve domain alignment, while SkillX~\cite{wang2026skillx} and MemP~\cite{fang2026memp} further leverage heterogeneous resources such as API documentation, execution traces, and external knowledge sources, reducing reliance on in-environment execution for skill acquisition.

Beyond initial skill construction, another line of work studies how skills can be iteratively improved through outcomes from execution. EvoSkill \cite{alzubi2026evoskill} and SkillForge \cite{liu2026skillforge} optimize skills through failure analysis: they collect failed execution trajectories, diagnose the underlying skill deficiencies, and rewrite the problematic portions to eliminate recurring failure modes. CoEvoSkills \cite{zhang2026coevoskills}, AutoRefine \cite{qiu2026autorefine}, and ProcMem \cite{mi2026procmem} adopt creation–evaluation–revision loops, where a skill generator and an independent verifier co-evolve through iterative cycles, using structured outcomes to progressively improve skill quality without ground-truth supervision. SkillClaw \cite{ma2026skillclaw} and Evolver \cite{wu2025evolver} take a collective approach, aggregating execution evidence across multiple users or agents to identify consistent success patterns and recurring failure modes, enabling cross-agent skill improvement where one user's experience benefits all others. SkillOpt \cite{yang2026skillopt} applies a deep-learning-style paradigm to iteratively refine skills—these approaches still lack a clear methodology for editing structured skill documents. From a reinforcement-learning perspective, SKILLRL and SAGE leverage the downstream task performance of skills as a reward signal to adjust the probability of trajectories that involve skill generation and utilization \cite{wang2026reinforcement, xia2026skillrl}. While these methods share the spirit of iterative text refinement, they predominantly operate at the level of individual prompts or whole-skill regeneration, lacking a systematic mechanism for making precise, traceable local modifications to structured skill documents with causal attribution of each edit's contribution.

\subsection{Post-Training RL Algorithms for LLMs}
Reinforcement learning (RL) has emerged as a core paradigm in modern LLM post-training, complementing supervised fine-tuning by improving reasoning capabilities and downstream task performance. Proximal Policy Optimization (PPO) \cite{schulman2017proximal} introduces clipped surrogate objectives and trust region constraints to stabilize policy updates. Direct Preference Optimization (DPO) \cite{rafailov2023direct} eliminates the reward model entirely by reparameterizing the RLHF objective into a classification loss over pairwise preference data. Group Relative Policy Optimization (GRPO) \cite{shao2024deepseekmath} eliminates the value network by computing group-relative advantages. Reinforcement Learning with Verifiable Rewards (RLVR) \cite{lambert2024tulu} further reduces dependence on human annotation by using automatically verifiable signals—such as code execution correctness or mathematical validity—as reward sources. Within this paradigm, several GRPO variants have been proposed to address its limitations. DAPO \cite{yu2026dapo} introduces decoupled advantage estimation to mitigate credit dilution across tokens of varying importance. Dr. GRPO \cite{liu2025understanding} identifies and corrects a bias in GRPO's group-level normalization that disproportionately penalizes longer rollouts. GSPO \cite{zheng2025group} proposes group sequence-level policy optimization to better align with sequence-level reward signals. SAPO \cite{gao2025soft} introduces a temperature-controlled soft gate mechanism that replaces hard clipping with smooth temperature-based attenuation, enabling more flexible control over policy deviation during training. While these post-training algorithms have proposed diverse modifications to GRPO across different training environments with promising results, their application to online advertising systems still necessitates further adaptations due to the unique characteristics of this domain.

\subsection{Delayed Outcome Modeling in Advertising}

Delayed feedback is common in advertising and recommendation,
where conversions or long-term outcomes become observable only
after a substantial lag \cite{zhang2021counterfactual}. One line
of research focuses on delayed conversion feedback, mitigating
immature or false-negative labels through feedback-shift
correction, multi-task learning, elapsed-time modeling, label
correction, debiased estimation, and continuous training
\cite{yasui2020feedback,hou2021conversion,yang2021capturing,
yasui2022learning,wang2023unbiased,chen2022asymptotically,
gu2021real}. DelayAdapter further formulates this problem as
unsupervised domain adaptation from reliably labeled historical
data to recent unlabeled traffic \cite{yu2024addressing}. These
methods primarily address delayed binary conversion labels. Another line of work uses early behavioral signals or short-term
proxies to estimate future outcomes. Post-click methods exploit
intermediate user behaviors to improve eventual conversion
prediction
\cite{su2021attention,yang2022generalized,guo2023leveraging},
while surrogate-based methods learn mappings from short-term
proxies to long-term outcomes \cite{athey2019surrogate}.
General-purpose forecasting models such as the Temporal Fusion
Transformer (TFT) can additionally model early multivariate
trajectories and static covariates \cite{lim2021temporal}.
Related studies also optimize recommendation policies under
delayed rewards through counterfactual reward modification or
delay-aware bandit learning
\cite{zhang2021counterfactual,mcdonald2023impatient}. Unlike these approaches, LRP predicts continuous long-term
rewards from short-term post-intervention signals, population
features, and intervention information. It explicitly models
population-invariant dynamics and population-specific deviations
while transferring relevant early response patterns from historical
experiments.
\section{Method}
\label{sec:method}

\begin{figure}[!tb]
  \centering
  \includegraphics[width=\linewidth]{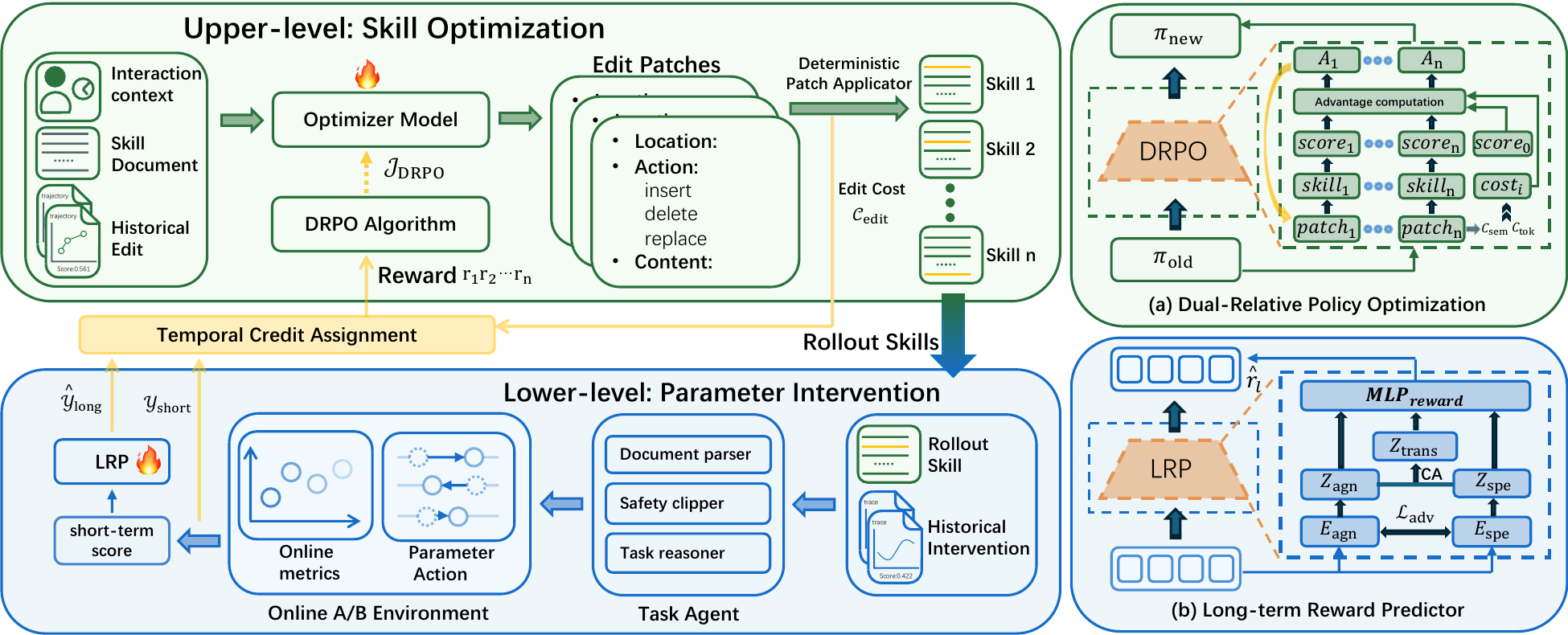}
  \caption{Overview of Document-Mediated Reinforcement Learning (DMRL) framework. The framework integrates the upper-level for skill optimization and lower-level for parameter intervention.}
  \label{fig:3_framework}
\end{figure}

As shown in Figure~\ref{fig:3_framework}, we propose Document-Mediated Reinforcement Learning (DMRL), a bilevel closed-loop framework for skill optimization in advertising systems. DMRL operates on two levels: an upper-level policy that iteratively edits structured skill documents, and a lower-level frozen task agent that interprets the modified documents to adjust advertising AB parameters in real time. The skill document serves as the semantic interface between the two levels. To obtain reliable advantage estimates in advertising system, we introduce Dual-Relative Policy Optimization (DRPO), which extends GRPO by handling reward outliers through robust within-group comparisons and leveraging treatment-control baselines for more reliable advantage estimation, while incorporating edit cost regularization for controlled and risk-aware skill evolution. To address the high latency of reward, the Long-term Reward Predictor (LRP) module estimates long-term rewards from short-term signals through disentangled population encoding and cross-attention historical transfer, enabling accurate prediction even for the underrepresented populations. To ensure stable training and reliable reward attribution, we employ a two-stage training strategy that first optimizes LRP for long-term reward estimation and then optimizes DRPO for policy learning, thereby improving the stability of both reward estimation and advantage computation.

\subsection{Long-term Reward Predictor}
In advertising systems, the long-term reward of a skill modification arrives with significant delay—often days after intervention—making it impractical to directly optimize the upper-level policy. To address this, we propose the Long-term Reward Predictor (LRP), a module that estimates long-term rewards from short-term signals by leveraging disentangled representation learning and cross-attention based transfer. Specifically, LRP explicitly disentangles population-invariant dynamics from population-specific deviations, and retrieves relevant historical patterns via cross-attention.

\subsubsection{Disentangled Representation Encoding}

To explicitly capture both population-invariant dynamics and population-specific deviations in the short-to-long reward mapping, we decompose the representation into two components: a population-agnostic representation and a population-specific representation.

First, a population-agnostic encoder $E_\mathrm{agn}$ takes the short-term signals $r_s$ and AB parameter modification $\Delta a$ as inputs, mapping the short-term signals and intervention signals into a population-invariant latent representation: 
\begin{align} 
z_{\mathrm{agn}} = E_{\mathrm{agn}}(r_s, \Delta a).
\end{align} 
Since population features $u$ are excluded from its input, $E_\mathrm{agn}$ is encouraged to model only the universal short-to-long dynamics that are common across populations.

In parallel, a population-specific encoder $E_{\mathrm{spe}}$ incorporates both the short-term signals,  intervention signals and the population features to capture group-dependent deviations: \begin{align} 
z_{\mathrm{spe}} = E_{\mathrm{spe}}(r_s, u, \Delta a). 
\end{align} 
This component is intended to represent the variation in short-to-long reward mapping across different user segments.

\subsubsection{Cross-Attention Historical Transfer}

To incorporate transferable historical dynamics from prior experiments, we introduce a memory bank equipped with a cross-attention retrieval mechanism.

Each historical experiment is encoded into a key-value pair:
\begin{align}
k_j &= E_{\mathrm{key}}(r_s^{(j)}, u^{(j)}, \Delta a^{(j)}), \\
v_j &= E_{\mathrm{value}}(\tau^{(j)}_{1d}),
\end{align}
where $r_s^{(j)}$ denotes the short-term signals observed within the first day, $u^{(j)}$ represents the user population features, $\Delta a^{(j)}$ is the corresponding parameter modification, and $\tau^{(j)}_{1d}$ denotes the multivariate within-day fluctuation trajectory of the $j$-th historical experiment. Intuitively, the key captures the coarse experimental condition for similarity matching, while the value stores fine-grained early response dynamics that is informative for long-term reward.

These historical records form a memory bank:
\begin{align}
\mathcal{M} = \{(k_j, v_j)\}_{j=1}^{M},
\end{align}
which is continuously updated as new experimental outcomes be-
come available during training.

For the current target instance, we construct the query representation by concatenating the disentangled features from the population-agnostic and population-specific branches:
\begin{align}
q = [z_{\mathrm{agn}} \,\|\, z_{\mathrm{spe}}].
\end{align}

We then retrieve relevant historical information through scaled dot-product attention over the memory bank:
\begin{align}
z_{\mathrm{trans}} = \mathrm{softmax}\left(\frac{qK^\top}{\sqrt{d_k}}\right)V,
\end{align}
where $K \in \mathbb{R}^{M \times d_k}$ and $V \in \mathbb{R}^{M \times d_v}$ denote the stacked key and value matrices, respectively. This retrieval process allows the model to identify historical experiments with similar short-term signals levels, user populations, and intervention settings, while aggregating their early dynamic response patterns as transferable evidence.

Finally, we fuse the population-agnostic, population-specific, and retrieved historical representations through a gating mechanism:
\begin{align}
z_{\mathrm{fused}} = z_{\mathrm{agn}} + z_{\mathrm{spe}} + \beta(q) \cdot z_{\mathrm{trans}},
\end{align}
where
\begin{align}
\beta(q) = \sigma(W_\beta q + b_\beta)
\end{align}
is an instance-wise gating function that controls the contribution of historical transfer. When the retrieved historical patterns are less relevant to the current instance, the gating value is reduced, allowing model to rely more on the encoded representations.

The fused representation is then decoded into the predicted long-term reward:
\begin{align}
\hat{r}_l = \mathrm{MLP}_{\mathrm{reward}}(z_{\mathrm{fused}}),
\end{align}
which serves as the estimated rollout feedback for the edit action.

\subsection{Dual-Relative Policy Optimization}

Directly applying standard GRPO to advertising recommendation suffers from three limitations. First, it is sensitive to volatile rewards, which can destabilize advantage estimation in online experiments. Second, it does not exploit the treatment-control structure of advertising experiments, where concurrent control-group outcomes offer a natural baseline. Third, it ignores the cost of modifying the skill itself when computing the optimization signal for edit actions. To address these limitations, we propose Dual-Relative Policy Optimization (DRPO), which combines robust within-group normalization, treatment-control baselines, and edit cost regularization to yield more reliable and risk-aware policy optimization.

\subsubsection{Edit Cost for Skill Modification}

In skill optimization, different edits induce substantially different levels of structural and semantic change, even when they yield similar estimated rewards. As shown in Figure~\ref{fig:3_framework}, each edit is represented as a structured tuple consisting of \emph{location}, \emph{action}, and \emph{content}. These components are not equally consequential: modifying different sections of a skill document affects different functional parts of the policy, different edit actions introduce different levels of disruption, and the edited content itself varies in both edit-level and semantic magnitude.

To capture this, we define the edit cost \(C_{\mathrm{edit}}(e)\) as a deterministic function of the structured edit:
\begin{align}
C_{\mathrm{edit}}(e)
=
w(\ell_e, a_e)
\left[
\eta \, C_{\mathrm{tok}}(e)
+
(1-\eta)\, C_{\mathrm{sem}}(e)
\right],
\end{align}
where \(\ell_e\) and \(a_e\) denote the edited location and action type, respectively, and \(w(\ell_e, a_e)\) is a structure-dependent weight that reflects the relative risk of editing different sections with different operations. The term \(C_{\mathrm{tok}}(e)\) measures the edit-level edit magnitude, while \(C_{\mathrm{sem}}(e)\) captures the semantic shift introduced by the edit.

Specifically, we define the edit-level edit magnitude as the normalized edit distance between the original and edited text:
\begin{align}
C_{\mathrm{tok}}(e)
=
\frac{\mathrm{EditDist}(x_{\mathrm{old}}, x_{\mathrm{new}})}
{\max\!\left(|x_{\mathrm{old}}|, |x_{\mathrm{new}}|\right)},
\end{align}
where \(x_{\mathrm{old}}\) and \(x_{\mathrm{new}}\) denote the original and edited text spans, respectively, and \(\mathrm{EditDist}(\cdot,\cdot)\) denotes the Levenshtein distance computed over token sequences.

We define the semantic shift as the cosine distance between the embeddings of the original and edited text:
\begin{align}
C_{\mathrm{sem}}(e)
=
1 - \cos\!\big(\phi(x_{\mathrm{old}}), \phi(x_{\mathrm{new}})\big),
\end{align}
where \(\phi(\cdot)\) denotes a sentence embedding model that maps text into a semantic representation space. A larger value of \(C_{\mathrm{sem}}(e)\) indicates a greater semantic deviation.

In this way, \(C_{\mathrm{tok}}(e)\) captures surface-form modification, while \(C_{\mathrm{sem}}(e)\) captures semantic deviation. Skill modifications that achieve similar reward gains but incur smaller structural or semantic disruption are preferred, thereby discouraging overly aggressive modifications and promoting more controlled skill evolution.

\subsubsection{Dual-Relative Advantage Estimation}

Following the rollout-based setting of GRPO, the optimizer generates \(G\) candidate atomic edits for each skill document, denoted by \(\{e_i\}_{i=1}^{G}\). Each edit is represented as \(e_i=(\ell_i, a_i, c_i)\), where \(\ell_i\), \(a_i\), and \(c_i\) denote the target location, edit action, and edited content, respectively. Each edited skill \(S_i\) is then executed by task agent, which instantiates the corresponding parameter intervention and evaluates it through online A/B experiment. In parallel, a concurrent control group runs under the best configuration and provides baseline outcome. Based on the resulting experimental signals, LRP produces the corresponding long-term reward \(R_i\) for rollout \(i\) and \(R_0\) for the control group, while each edit \(e_i\) is associated with an edit cost term \(C_{\mathrm{edit}}(e_i)\).

We incorporate robust within-group comparison, treatment-control comparison, and edit cost into advantage estimation:

\begin{align}
A_i
&=
\underbrace{
\alpha
\frac{R_i - \tilde{\mu}}
{1.4826 \cdot \mathrm{MAD} + \delta}
}_{\text{Term 1}}
+
\underbrace{
(1-\alpha)\frac{R_i - R_{0}}{R_{0} + \delta}
}_{\text{Term 2}}
-
\underbrace{
\lambda_{\mathrm{edit}}C_{\mathrm{edit}}(e_i)
}_{\text{Term 3}}.
\end{align}

Here,
\begin{align}
\tilde{\mu} &= \mathrm{median}(R_i),\\
\mathrm{MAD} &= \mathrm{median}\left(|R_i-\tilde{\mu}|\right).
\end{align}

Term 1 provides a robust within-group comparison by replacing the mean and standard deviation used in standard normalization with the median and median absolute deviation (MAD), which reduces sensitivity to outlier rollouts. The factor \(1.4826\) rescales MAD to a standard-deviation-consistent estimator under normality, preserving interpretability of the normalized score.

Term 2 measures the improvement of each rollout relative to the control group, grounding the advantage in the treatment-control structure. The coefficient \(\alpha \in [0,1]\) balances within-group comparison and control-group-relative improvement.

Term 3 penalizes the cost of modifying the skill document. By explicitly incorporating edit cost into the utility, this term encourages the policy to prefer effective yet minimally disruptive revisions, promoting controlled and risk-aware skill evolution. 

\subsection{Two-Stage Training Strategy}

To ensure stable and effective population-aware long-term reward prediction and policy optimization, we adopt a two-stage training strategy. In Stage I, we train the Long-term Reward Predictor (LRP) to model delayed rewards under population heterogeneity. In Stage II, we freeze LRP and optimize the skill-editing policy using DRPO.

\subsubsection{Stage I: Training the Long-Term Reward Predictor}

In Stage I, we train LRP to predict long-term rewards from
short-term signals, population features, and intervention
information. In addition to minimizing the reward prediction error,
we employ adversarial regularization to encourage
$z_{\mathrm{agn}}$ to capture population-invariant dynamics.
Specifically, a population discriminator $D_{\mathrm{adv}}$
attempts to infer the population label from $z_{\mathrm{agn}}$,
while the population-agnostic encoder $E_{\mathrm{agn}}$ is
optimized to make such inference difficult.

As shown in Algorithm ~\ref{alg:lrp_training}, we adopt a three-phase training schedule, to improve training stability. First, we warm up the reward-prediction network with $\lambda_{\mathrm{adv}}=0$, allowing the encoders and reward predictor to learn representations informative for long-term reward estimation. Second, we freeze $E_{\mathrm{agn}}$ and pretrain $D_{\mathrm{adv}}$ to predict population labels from $z_{\mathrm{agn}}$. Finally, we jointly optimize LRP and the
population discriminator using a gradient reversal layer (GRL)
\cite{ganin2015unsupervised}.

Given the population-agnostic representation
\begin{align}
z_{\mathrm{agn}}
=
E_{\mathrm{agn}}(r_s,\Delta a),
\end{align}
the discriminator predicts the population label as
\begin{align}
\hat{g}
=
D_{\mathrm{adv}}
\left(
\operatorname{GRL}(z_{\mathrm{agn}})
\right).
\end{align}
The GRL acts as an identity mapping during the forward pass, but
reverses the gradient propagated to $E_{\mathrm{agn}}$ during
backpropagation:
\begin{align}
\operatorname{GRL}(z)=z,
\qquad
\frac{\partial \operatorname{GRL}(z)}{\partial z}
=-\mathbf{I}.
\end{align}

The population classification loss is defined as
\begin{align}
\mathcal{L}_{\mathrm{adv}}
=
-\mathbb{E}
\left[
\log
D_{\mathrm{adv}}
\left(
g \mid \operatorname{GRL}(z_{\mathrm{agn}})
\right)
\right],
\end{align}
where $g$ denotes the ground-truth population label. The overall
training objective of LRP is
\begin{align}
\mathcal{L}_{\mathrm{LRP}}
=
\mathrm{MSE}(\hat{r}_l,r_l)
+
\lambda_{\mathrm{adv}}\mathcal{L}_{\mathrm{adv}},
\end{align}
where $\hat{r}_l$ and $r_l$ denote the predicted and observed
long-term rewards, respectively, and $\lambda_{\mathrm{adv}}$
controls the strength of adversarial regularization. During joint
training, minimizing $\mathcal{L}_{\mathrm{adv}}$ improves the
population-classification ability of $D_{\mathrm{adv}}$, whereas
the gradient reversal operation drives $E_{\mathrm{agn}}$ in the
opposite direction, thereby reducing population-identifying
information in $z_{\mathrm{agn}}$.

\begin{algorithm}[t]
\caption{Three-Phase Training of LRP}
\label{alg:lrp_training}
\small
\begin{algorithmic}[1]
\Require Dataset $\mathcal{D}$, memory bank $\mathcal{M}$,
adversarial weight $\lambda_{\mathrm{adv}}$
\Ensure Trained $\Theta_{\mathrm{LRP}}$

\State $\Theta_{\mathrm{LRP}}
\gets
\{\theta_{\mathrm{agn}},\theta_{\mathrm{spe}},
\theta_{\beta},\theta_{\mathrm{reward}}\}$

\State \textbf{Phase I: Reward-prediction warm-up}
\For{each iteration}
    \State $\mathcal{B}\sim\mathcal{D}$
    \State $\mathcal{B},\mathcal{M}
    \xrightarrow{\mathrm{LRP}}
    \hat r_l$
    \State $\mathcal{L}_{\mathrm{pred}}
    \gets \mathrm{MSE}(\hat r_l,r_l)$
    \State $\Theta_{\mathrm{LRP}}
    \gets
    \Theta_{\mathrm{LRP}}
    -\eta_{\mathrm{lrp}}
    \nabla_{\Theta_{\mathrm{LRP}}}
    \mathcal{L}_{\mathrm{pred}}$
\EndFor

\State \textbf{Phase II: Discriminator warm-up}
\State Freeze $\theta_{\mathrm{agn}}$
\For{each iteration}
    \State $\mathcal{B}\sim\mathcal{D}$
    \State $(r_s,\Delta a)
    \xrightarrow{E_{\mathrm{agn}}}
    z_{\mathrm{agn}}
    \xrightarrow{D_{\mathrm{adv}}}
    \hat g$
    \State $\mathcal{L}_{\mathrm{adv}}
    \gets -\mathbb{E}[\log D_{\mathrm{adv}}
    (g\mid z_{\mathrm{agn}})]$
    \State $\theta_{\mathrm{adv}}
    \gets
    \theta_{\mathrm{adv}}
    -\eta_{\mathrm{adv}}
    \nabla_{\theta_{\mathrm{adv}}}
    \mathcal{L}_{\mathrm{adv}}$
\EndFor

\State \textbf{Phase III: Joint adversarial training}
\State Unfreeze $\theta_{\mathrm{agn}}$
\For{each iteration}
    \State $\mathcal{B}\sim\mathcal{D}$
    \State $\mathcal{B},\mathcal{M}
    \xrightarrow{\mathrm{LRP}}
    (\hat r_l,z_{\mathrm{agn}})$
    \State $z_{\mathrm{agn}}
    \xrightarrow{\mathrm{GRL}}
    D_{\mathrm{adv}}
    \rightarrow \hat g$
    \State $\mathcal{L}_{\mathrm{LRP}}
    \gets
    \mathcal{L}_{\mathrm{pred}}
    +\lambda_{\mathrm{adv}}\mathcal{L}_{\mathrm{adv}}$
    \State $(\Theta_{\mathrm{LRP}},\theta_{\mathrm{adv}})
    \gets
    (\Theta_{\mathrm{LRP}},\theta_{\mathrm{adv}})
    -(\eta_{\mathrm{lrp}},\eta_{\mathrm{adv}})
    \nabla\mathcal{L}_{\mathrm{LRP}}$
\EndFor

\State \Return $\Theta_{\mathrm{LRP}}$
\end{algorithmic}
\end{algorithm}

\subsubsection{Stage II: Optimizing the Skill-Editing Policy with DRPO}

After LRP is trained, we freeze its parameters and use it to provide reward estimates for policy learning.

The policy is optimized using the clipped surrogate objective
\begin{align}
\mathcal{L}_{\mathrm{DRPO}}(\theta)
&=
- \mathbb{E}_{i,t}
\Bigl[
\min\Bigl(
\rho_{i,t}(\theta) A_i,\,
\operatorname{clip}\bigl(\rho_{i,t}(\theta), 1-\epsilon_c, 1+\epsilon_c\bigr) A_i
\Bigr)
\Bigr]
\nonumber 
+ \beta \, D_{\mathrm{KL}}\!\bigl(\pi_{\theta} \,\Vert\, \pi_{\theta_{\mathrm{old}}}\bigr),
\end{align}
where
\begin{align}
\rho_{i,t}(\theta)
=
\frac{\pi_{\theta}\bigl(e_{i,t} \mid e_{i,<t}\bigr)}
{\pi_{\theta_{\mathrm{old}}}\bigl(e_{i,t} \mid e_{i,<t}\bigr)}.
\end{align}
Here, \(e_{i,t}\) denotes the \(t\)-th token in the serialized representation of edit \(e_i\), and \(e_{i,<t}\) denotes its prefix. The clipped ratio stabilizes policy updates, while the KL regularization term constrains the policy from drifting excessively from the previous iteration.

\section{Experiments}
\label{sec:experiments}

\section{Experiment}

To validate the effectiveness of our DMRL framework, we conduct a series of experiments to answer the following research questions:

RQ1: How effective is DMRL for skill optimization in advertising recommendation, compared with existing skill optimization methods and long-term reward modeling approaches?

RQ2: How do the key components of DMRL, including LRP, DRPO, and the two-stage training strategy, individually and jointly contribute to the overall performance of the framework?

RQ3: How sensitive is DMRL to implementation such as the number of DRPO rollouts and the scale of backbone model?

\subsection{Experimental Setup}

\subsubsection{Data}

According to the heterogeneous short-to-long mappings observed in Figure~\ref{fig:1_population}, we partition users into four predefined population groups: Low-ST Low-LT, High-ST Low-LT, Low-ST High-LT, and High-ST High-LT. Specifically, Low-ST Low-LT denotes the users with low short-term and low long-term metrics, while the other groups are defined analogously. The population labels are determined using historical behavioral statistics and remain fixed throughout the experiment. Following this partition, we adopt a population-aware data construction strategy. Specifically, we separately collect eligible trajectories from each population group and combine them to construct the LRP dataset. As shown in Table~\ref{tab:population-distribution}, the resulting dataset covers all four population groups while approximately preserving their distribution in online traffic. Each sample corresponds to a user trajectory associated with a parameter intervention and contains the parameter changes, user features, short-term feedback, population label, and observed long-term outcome. The resulting dataset contains approximately 12M trajectories, obtained by low-rate sampling from a massive online traffic pool. The sampling procedure is applied consistently across population groups and approximately preserves their relative prevalence in online traffic. Because LRP is trained on historical experiments and subsequently applied to future online interventions, we adopt a chronological split to approximate the deployment setting. Specifically, trajectories are ordered by intervention time and divided into training, validation, and test sets. For each target trajectory, the memory bank contains only historical records whose long-term outcomes have fully matured before the target intervention, preventing temporal information leakage.

\subsubsection{Implementation Details}

DMRL is implemented as a hierarchical skill-optimization framework comprising upper-level skill optimization and lower-level task execution.

At the upper level, an optimizer model iteratively improves its skill-optimization strategy using feedback collected from previous rollouts. At each iteration, the optimizer proposes modifications to the structured skills, which are subsequently instantiated and executed by a frozen downstream task agent in the online experimentation environment. The environment returns short-term outcomes and treatment--control comparison signals, while LRP estimates the corresponding long-term rewards. These signals are incorporated into DRPO to update the skill-optimization strategy, thereby establishing a closed-loop skill-optimization pipeline under delayed rewards. Unless otherwise specified, we instantiate the optimizer with Qwen3-8B and sample 8 rollouts in DRPO iteration.

At the lower level, we employ OpenAI Codex as the frozen Task Agent, using GPT-5.5 as its underlying model. After experimental round \(t\), the upper-level optimizer produces an updated skill \(S_{t+1,i}\) for the \(i\)-th rollout. After loading \(S_{t+1,i}\), the Task Agent generates a parameter configuration for round \(t+1\) based on the current experimental state and applies the configuration through tool calls. To mitigate operational risks, the Task Agent is restricted to a predefined whitelist of parameters and permissible value ranges.

\begin{table}[t]
\centering
\caption{Comparison of population distributions.}
\label{tab:population-distribution}

\begin{minipage}{0.48\linewidth}
\centering
\begin{tabular}{@{}lc@{}}
\toprule
\multicolumn{2}{c}{\textbf{(a) Online traffic distribution}} \\
\midrule
Population & Proportion \\
\midrule
Low-ST Low-LT   & 41.11\% \\
High-ST Low-LT  & 25.85\% \\
Low-ST High-LT  & 13.79\% \\
High-ST High-LT & 19.25\% \\
\bottomrule
\end{tabular}
\end{minipage}
\hfill
\begin{minipage}{0.48\linewidth}
\centering
\begin{tabular}{@{}lc@{}}
\toprule
\multicolumn{2}{c}{\textbf{(b) Dataset statistics}} \\
\midrule
Population & Number \\
\midrule
Low-ST Low-LT   & 4,933,847 \\
High-ST Low-LT  & 3,102,196\\
Low-ST High-LT  & 1,655,732\\
High-ST High-LT & 2,310,468 \\
\bottomrule
\end{tabular}
\end{minipage}
\end{table}

\subsubsection{Evaluation Metrics}
We evaluate the online effectiveness of DMRL using two business-oriented metrics and one composite metric: App Usage Duration (AUD), Posterior Expected Spend (PES), and Life Time Value (LTV), which capture user engagement, advertising monetization effectiveness, and overall platform economic value, respectively. To measure both immediate and delayed effects, we report each base metric over two distinct time horizons. Specifically, short-term metrics include AUD@1d and PES@1d, measured within one day after intervention, while long-term metrics include AUD@7d, PES@7d and LTV, measured over a 7-day horizon.

To ensure the reliability, we adopt several stabilization strategies. First, 7-day cumulative metric mitigates the impact of short-term fluctuations. Second, we apply CUPED \cite{deng2013improving} during online evaluation, using user-level historical statistics from the pre-experiment A/A period as covariates, to reduce the variance caused by users' intrinsic behavioral. Importantly, experiments in different tables are conducted during non-overlapping time windows, involving different traffic allocations. Within each table, all methods are evaluated under same eligibility criteria, traffic distribution and observation window. All conclusions are restricted to within-table comparisons.

\noindent\textbf{App Usage Duration (AUD).}
AUD measures the amount of time users spend in the application after intervention and serves as a proxy for user engagement. A higher AUD value generally indicates stronger engagement with the recommended content.

\noindent\textbf{Posterior Expected Spend (PES).}
PES reflects the monetization of the advertising business, the platform's primary source of revenue. By calibrating the original expected spend with post-hoc signals, such as conversion performance and actual cost, PES provides a robust estimate of the revenue impact from advertising monetization.

\noindent\textbf{Life Time Value (LTV).}
LTV converts per-metric gains and losses into economic equivalents and aggregates them into a comprehensive economic assessment that accounts for user engagement and monetization contributions. The monetization component captures the platform’s overall monetization efficiency, including advertising. We regard LTV improvement greater than 0.01\% as a practically meaningful gain
for evaluating the effectiveness of the intervention.

\begin{table}[t]
\centering
\caption{Comparison of DMRL with state-of-the-art methods in skill optimization and long-term reward modeling.}
\label{tab:sota_comparison}
\resizebox{\linewidth}{!}{
\begin{tabular}{lccccc}
\toprule
Method & AUD@1d & PES@1d & AUD@7d & PES@7d & LTV \\
\midrule
\multicolumn{6}{l}{\textit{Skill Optimization Methods}} \\
LRP+SAGE \cite{wang2026reinforcement} & +0.049\% & +0.757\% & -0.094\% & -0.911\% & -0.049\% \\
LRP+SkillOpt \cite{yang2026skillopt} & +0.012\% & -0.515\% & +0.069\% & +0.548\% & +0.024\% \\
LRP+SKILLRL \cite{xia2026skillrl} & +0.101\% & -1.547\% & +0.133\% & -0.430\% & +0.004\% \\
\midrule
\multicolumn{6}{l}{\textit{Long-term Reward Modeling Methods}} \\
DRPO+TFT \cite{lim2021temporal} & -0.187\% & -0.085\% & +0.018\% & +0.656\% & +0.015\% \\
DRPO+DelayAdapter \cite{yu2024addressing} & -0.158\% & +1.480\% & -0.060\% & +0.134\% & +0.005\% \\
\midrule
\multicolumn{6}{l}{\textit{Our Framework}} \\
DMRL (DRPO+LRP) & -0.020\% & +0.644\% & +0.082\% & +0.960\% & +0.052\% \\
\bottomrule
\end{tabular}
}
\end{table}

\subsection{Main Result: RQ1}

We employ two groups of baselines: skill optimization methods and long-term reward modeling methods. To isolate each module's contribution, we keep LRP fixed when comparing DRPO with skill optimization baselines, and keep DRPO fixed when comparing LRP with long-term reward modeling baselines. As summarized in Table ~\ref{tab:sota_comparison}, DMRL exhibits a substantial and consistent improvement over other methods, with the highest LTV improvement of \(+0.052\%\).

Compared with skill optimization baselines, DRPO outperforms SAGE, SkillOpt, and SKILLRL by \(+0.101\%\), \(+0.028\%\), and \(+0.048\%\). Specifically, SAGE brings positive short-term metrics, but it degrades both long-term AUD and PES, resulting in a negative LTV of \(-0.049\%\). SkillOpt achieves competitive monetization improvements on AUD@7d (\(+0.069\%\)) and PES@7d (\(+0.548\%\)), resulting in better LTV performance (\(+0.024\%\)) than SAGE. SKILLRL further improves user engagement on both AUD@1d (\(+0.101\%\)) and AUD@7d (\(+0.133\%\)), but its monetization metrics drop significantly, with PES@1d and PES@7d decreasing by \(-1.547\%\) and \(-0.430\%\), respectively. In contrast, DRPO achieves a more balanced trade-off. Although AUD@1d slightly decreases by \(-0.020\%\), positive gains on PES@1d (\(+0.644\%\)), AUD@7d (\(+0.082\%\)), and PES@7d (\(+0.960\%\)) demonstrate that skill optimization should account for engagement and monetization outcomes across multiple temporal horizons. By integrating these feedback signals into policy optimization, DMRL supports stable skill updates and achieves a high long-term value. 

For long-term reward modeling methods, TFT and DelayAdapter achieve positive LTV improvements, suggesting that explicitly modeling delayed rewards can provide useful optimization signals. However, their engagement-side performance remains relatively weak. Specifically, TFT improves PES@7d by \(+0.656\%\), indicating its ability to capture delayed monetization signals, but it suffers from short-term degradation on AUD@1d (\(-0.187\%\)) and PES@1d (\(-0.085\%\)), resulting in a moderate LTV improvement of \(+0.015\%\). DelayAdapter further improves PES@1d to \(+1.480\%\), but it degrades all long-term metrics, yielding only a marginal LTV gain of \(+0.005\%\). In contrast, LRP achieves a more balanced improvement and outperforms TFT and DelayAdapter in LTV by \(+0.037\%\) and \(+0.047\%\), respectively. These results demonstrate that LRP better balances monetization, engagement, and delayed economic value. By capturing population-specific preferences through user-group feature extraction and historical pattern transfer, LRP provides more reliable long-term reward signals for policy optimization.

Taken together, DMRL combines the strengths of LRP and DRPO: LRP estimates population-aware delayed rewards, while DRPO performs risk-aware and edit-regularized policy optimization. This joint design enables DMRL to achieve the best aggregate outcome.

\begin{table}[t]
\centering
\caption{Ablation study of variants of LRP. UE denotes unified encoder and MB denotes memory bank module.}
\label{tab:lrp_ablation}
\resizebox{\linewidth}{!}{
\begin{tabular}{lccccc}
\toprule
Method & AUD@1d & PES@1d & AUD@7d & PES@7d & LTV \\
\midrule
UE & +0.064\% & +0.098\% & +0.061\% & +0.257\% & +0.031\% \\
w/o MB & -0.062\% & +0.409\% & -0.010\% & +0.491\% & +0.020\% \\
UE w/o MB & -0.011\% & -0.245\% & +0.025\% & -0.231\% & -0.028\% \\
LRP  & +0.006\% & +0.674\% & +0.155\% & +0.662\% & +0.042\% \\
\bottomrule
\end{tabular}
}
\end{table}

\begin{table}[t]
\centering
\caption{Ablation study of DRPO components. MN, RR, and EC denote MAD-based normalization, reference reward relative to the control group, and edit cost, respectively.}
\label{tab:drpo_ablation}
\resizebox{\linewidth}{!}{
\begin{tabular}{lccccc}
\toprule
Method & AUD@1d & PES@1d & AUD@7d & PES@7d & LTV \\
\midrule
w/o MN & +0.067\% & -0.849\% & +0.069\% & -1.177\% & +0.008\% \\
w/o RR & +0.194\% & +1.411\% & +0.125\% & -0.371\% & +0.013\% \\
w/o EC & +0.100\% & +0.070\% & -0.007\% & -0.688\% & -0.018\% \\
GRPO & +0.051\% & +0.162\% & -0.097\% & -1.216\% & -0.069\% \\
DRPO & -0.048\% & +1.197\% & +0.131\% & +0.059\% & +0.021\% \\
\bottomrule
\end{tabular}
}
\end{table}

\subsection{More Result: RQ2}

\subsubsection{LRP}

To test the main design in LRP, we conduct an ablation study over different variants. As shown in Table ~\ref{tab:lrp_ablation}, the full LRP model achieves the best overall LTV improvement, while obtaining the strongest gains on PES@1d, AUD@7d, and PES@7d.

The UE variant achieves the strongest short-term engagement gain on AUD@1d (\(+0.064\%\)) and obtains a competitive LTV of \(+0.031\%\), with PES@1d and PES@7d improving by only \(+0.098\%\) and \(+0.257\%\), respectively. These results suggest that relying solely on unified representations is insufficient to fully capture heterogeneous short-to-long reward mappings across user populations. The variant without the memory bank achieves positive monetization gains on PES@1d (\(+0.409\%\)) and PES@7d (\(+0.491\%\)), but it degrades both AUD@1d and AUD@7d, resulting in a lower LTV of \(+0.020\%\). This indicates that the memory bank helps calibrate delayed reward estimates and prevent the model from overemphasizing monetization signals at the expense of user engagement. The UE w/o MB variant performs the worst among all variants with negative gains on AUD@1d, PES@1d, PES@7d, and LTV, further demonstrating the complementary benefits of population-aware representation learning and memory-based historical transfer. 

Overall, the ablation results show that different LRP components contribute to complementary aspects of performance. Together, these components produce a more balanced and economically aligned reward estimate for downstream DRPO optimization.

\subsubsection{DRPO}

To study the contribution of the three key modifications introduced in DRPO, we conduct a systematic ablation study. As shown in Table ~\ref{tab:drpo_ablation}, DRPO achieves the highest LTV of \(+0.021\%\) among all variants, with positive gains on PES@1d (\(+1.197\%\)), AUD@7d (\(+0.131\%\)), and PES@7d (\(+0.059\%\)).

The w/o MN variant improves AUD@1d and AUD@7d by \(+0.067\%\) and \(+0.069\%\), respectively, but its PES@1d and PES@7d decrease by \(-0.849\%\) and \(-1.177\%\). As a result, its LTV improvement is limited to \(+0.008\%\), suggesting that without robust normalization, the optimizer is more vulnerable to heavy-tailed reward signals. The w/o RR variant achieves the strongest short-term gains, improving AUD@1d and PES@1d by \(+0.194\%\) and \(+1.411\%\), respectively. However, its PES@7d decreases by \(-0.371\%\), and the resulting LTV improvement is only \(+0.013\%\). These results indicate that removing the control-group-relative reward reference makes policy optimization overemphasize immediate responses while weakening its alignment with delayed monetization effects. The w/o EC variant yields negative long-term metric changes and produces a negative LTV change of \(-0.018\%\). Similar to the w/o EC variant, vanilla GRPO obtains small short-term gains on AUD@1d (\(+0.051\%\)) and PES@1d (\(+0.162\%\)); however, it substantially hurts AUD@7d and PES@7d by \(-0.097\%\) and \(-1.216\%\), respectively, leading to the lowest LTV change of \(-0.069\%\). This highlights the limitation of directly applying the original GRPO objective to delayed reward optimization.

Collectively, the ablation results support the design rationale of DRPO. Together, these components provide more reliable advantage estimates, enabling DRPO to better balance short-term feedback, delayed outcomes, and aggregate economic value.

\begin{table}[t]
\centering
\caption{Comparison of different training strategies of DMRL.}
\label{tab:training_strategy}
\resizebox{\linewidth}{!}{
\begin{tabular}{lccccc}
\toprule
Strategy  & AUD@1d & PES@1d & AUD@7d & PES@7d & LTV \\
\midrule
Single-stage & -0.013\% & -0.424\% & +0.026\% & -0.702\% & -0.039\% \\
Two-stage & -0.020\% & +0.644\% & +0.082\% & +0.960\% & +0.052\% \\
\bottomrule
\end{tabular}
}
\end{table}

\subsubsection{Two-stage Training}

Finally, we examine the necessity of the two-stage training strategy adopted in DMRL, where LRP is first trained to predict delayed rewards and is then used to guide downstream policy optimization. As shown in Table ~\ref{tab:training_strategy}, we compare this design with direct single-stage optimization.

The single-stage strategy achieves a slightly better AUD@1d than the two-stage strategy, with a smaller decrease of \(-0.013\%\) compared with \(-0.020\%\). However, it leads to clear degradation in monetization-related metrics, reducing in PES@1d and PES@7d by \(-0.424\%\) and \(-0.702\%\), respectively, and resulting in a negative LTV change of \(-0.039\%\). In contrast, the two-stage strategy consistently improves long-term and monetization metrics, increasing AUD@7d by \(+0.082\%\), PES@1d by \(+0.644\%\), and PES@7d by \(+0.960\%\). Overall, the two-stage strategy achieves a positive LTV gain of \(+0.052\%\). These results suggest that jointly optimizing reward prediction and policy learning in a single stage causes LRP to overemphasize short-term signals and provide unstable supervision for policy updates. By first training LRP on historical trajectories, the two-stage strategy produces a more reliable proxy reward, thereby better guiding DRPO toward long-term value optimization.

\begin{table}[t]
\centering
\caption{Comparison of different Qwen3 backbones.}
\label{tab:qwen3_backbone}
\resizebox{\linewidth}{!}{
\begin{tabular}{lccccc}
\toprule
Backbone & AUD@1d & PES@1d & AUD@7d & PES@7d & LTV \\
\midrule
Qwen3-0.6B & -0.160\% & +1.572\% & +0.137\% & -0.606\% & -0.021\% \\
Qwen3-1.7B & +0.018\% & +0.303\% & +0.083\% & -0.341\% & +0.009\% \\
Qwen3-4B & -0.143\% & +1.264\% & +0.046\% & -0.291\% & +0.012\% \\
Qwen3-8B & -0.060\% & +1.135\% & +0.025\% & +0.539\% & +0.064\% \\
Qwen3-14B & -0.001\% & +0.806\% & +0.091\% & -0.529\% & +0.030\% \\
\bottomrule
\end{tabular}
}
\end{table}

\begin{table}[t]
\centering
\caption{Ablation study on the rollout number for Qwen3-8B.}
\label{tab:rollout_number}
\resizebox{\linewidth}{!}{
\begin{tabular}{lccccc}
\toprule
\#Rollouts  & AUD@1d & PES@1d & AUD@7d & PES@7d & LTV \\
\midrule
4 & -0.026\% & +0.081\% & -0.006\% & -0.169\% & -0.003\% \\
6 & +0.128\% & -0.639\% & +0.129\% & -1.219\% & +0.008\% \\
8 & +0.045\% & +0.307\% & +0.063\% & +1.095\% & +0.050\% \\
10 & -0.103\% & +0.128\% & +0.067\% & +0.064\% & +0.037\% \\
\bottomrule
\end{tabular}
}
\end{table}

\subsection{Ablation Study: RQ3}

\subsubsection{Ablation on Qwen3 Backbone Scale}
As shown in Table~\ref{tab:qwen3_backbone}, Qwen3-8B achieves the best overall performance among all evaluated backbones, yielding the highest LTV improvement of \(+0.064\%\) and positive gains on PES@1d (\(+1.135\%\)), AUD@7d (\(+0.025\%\)), and PES@7d (\(+0.539\%\)). Qwen3-0.6B achieves a notable PES@1d improvement of \(+1.572\%\), but it suffers from a substantial PES@7d degradation of \(-0.606\%\) and results in a negative LTV change of \(-0.021\%\). Qwen3-1.7B obtains the largest AUD@1d gain (\(+0.018\%\)) and achieves positive improvements on both PES@1d and AUD@7d. However, its PES@7d remains negative, leading to only a marginal LTV improvement of \(+0.009\%\). Qwen3-4B performs slightly better in terms of LTV (\(+0.012\%\)), but its improvements on long-term engagement are limited and PES@7d still decreases by \(-0.291\%\). These results suggest that smaller backbones lack sufficient capacity to generate high-quality candidate edits, leaving the optimization space underexplored. Increasing the backbone size to 14B does not lead to monotonic improvements. Qwen3-14B obtains negative PES@7d \(-0.529\%\) and weak PES@1d (\(+0.806\%\)), resulting in a lower LTV gain (\(+0.030\%\)) than Qwen3-8B, which suggests that increased model capacity introduces more diverse or aggressive edits that do not necessarily improve long-term skill optimization.

\subsubsection{Number of Rollouts in DRPO}
As shown in Table~\ref{tab:rollout_number}, the number of rollouts has a clear non-monotonic effect on online performance. Among all settings, 8 rollouts achieve the highest LTV improvement of \(+0.050\%\) and yields consistent gains across all metrics, including AUD@1d (\(+0.045\%\)), PES@1d (\(+0.307\%\)), AUD@7d (\(+0.063\%\)), and PES@7d (\(+1.095\%\)). With only 4 rollouts, the model obtains a slight PES@1d improvement of \(+0.081\%\), but AUD@1d, AUD@7d, and PES@7d all decrease, resulting in a negative LTV change of \(-0.003\%\). Increasing the rollout number to 6 improves engagement-related metrics, with AUD@1d and AUD@7d increasing by \(+0.128\%\) and \(+0.129\%\). However, reducing PES@1d and PES@7d lead to marginal LTV gain \(+0.008\%\). This suggests that insufficient rollout diversity limits the optimizer's ability to explore informative candidate edits. Further increasing the number of rollouts to 10 also fails to bring additional benefits. 10 rollouts maintain weaker gains on AUD@7d (\(+0.067\%\)), PES@1d (\(+0.128\%\)), and PES@7d (\(+0.064\%\)) than those achieves with 8 rollouts, and AUD@1d drops by \(-0.103\%\). As a result, the LTV improvement decreases from \(+0.050\%\) to \(+0.037\%\). These observations suggest that excessive rollouts introduce redundant candidate edits, which can dilute the effectiveness of reward-based selection.

\subsubsection{Qualitative Analysis of LRP}

\begin{figure}[h]
  \centering
  \includegraphics[width=\linewidth]{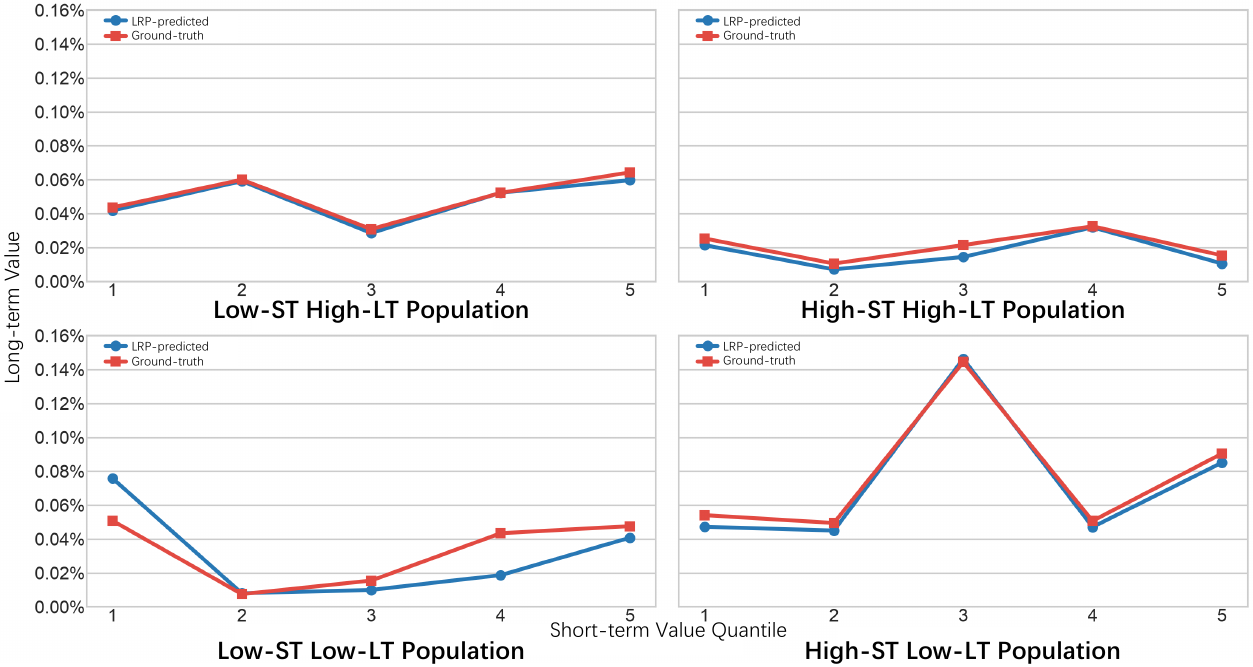}
  \caption{Comparison between LRP-predicted and Ground-truth LTV at short-term value quantiles across populations.}
  \label{fig:4_LRP}
\end{figure}

To qualitatively examine LRP's modeling behavior, we sample instances from different populations and visualize the distribution of LRP-predicted and ground-truth values in Figure~\ref{fig:4_LRP}. In all populations, the predicted values preserve the ordering of all the ground-truth observations even when the relationship is non-monotonic. For the Low-ST High-LT population, the two curves are closely aligned across all quantiles, with a mean absolute error of only \(0.0019\%\). LRP slightly underestimates the ground truth value at most quantiles, with the largest deviation occurring at the fifth quantile \(0.0046\%\). A similar pattern is observed for the High-ST High-LT population: LRP reproduces the overall trajectory but exhibits a small systematic underestimation, with a mean absolute error of \(0.0039\%\). For the High-ST Low-LT population, LRP successfully captures the pronounced peak at the third quantile and the subsequent decline, achieving a mean absolute error of \(0.0044\%\). The largest discrepancy occurs in the Low-ST Low-LT population, where LRP overestimates the first quantile by \(0.0250\%\) and underestimates the fourth quantile by \(0.0247\%\). Nevertheless, it still captures the overall population-specific trend. These results provide evidence that LRP can capture the heterogeneous short-to-long-term relationships across populations.
\section{Conclusion}
\label{sec:conclusion}

This paper proposes DMRL, a skill optimization framework in advertising recommendation under noisy, delayed, and heterogeneous feedback. DMRL decomposes the pipeline into robust optimizer training and long-term reward modeling, enabling the system to optimize edit decisions beyond immediate responses and toward long-term outcomes. Through the integration of DRPO for robust optimizer training under delayed feedback and LRP for long-term value modeling under population heterogeneity, the framework achieves reliable and effective optimization, while a two-stage training strategy further stabilizes the optimization and reward modeling process. Deployed in a large-scale advertising recommendation system, DMRL demonstrates significant improvements over baselines, validating its practical utility in real-world industrial environments. Looking ahead, this work opens several avenues for future research, including extending automated skill optimization to richer multi-objective scenarios and improving population-aware long-term reward modeling under evolving traffic distributions. We believe DMRL not only provides a deployable solution for automatic skill optimization, but also lays the groundwork for long-horizon optimization in dynamic recommendation systems.


\bibliographystyle{plainnat}
\bibliography{references}

@String{Chelsea = "Chelsea" }

@article{yao2022react,
  title={React: Synergizing reasoning and acting in language models},
  author={Yao, Shunyu and Zhao, Jeffrey and Yu, Dian and Du, Nan and Shafran, Izhak and Narasimhan, Karthik and Cao, Yuan},
  journal={arXiv preprint arXiv:2210.03629},
  year={2022}
}

@article{wang2023voyager,
  title={Voyager: An open-ended embodied agent with large language models},
  author={Wang, Guanzhi and Xie, Yuqi and Jiang, Yunfan and Mandlekar, Ajay and Xiao, Chaowei and Zhu, Yuke and Fan, Linxi and Anandkumar, Anima},
  journal={arXiv preprint arXiv:2305.16291},
  year={2023}
}

@inproceedings{zeng2024agenttuning,
  title={Agenttuning: Enabling generalized agent abilities for llms},
  author={Zeng, Aohan and Liu, Mingdao and Lu, Rui and Wang, Bowen and Liu, Xiao and Dong, Yuxiao and Tang, Jie},
  booktitle={Findings of the Association for Computational Linguistics: ACL 2024},
  pages={3053--3077},
  year={2024}
}

@article{shao2024deepseekmath,
  title={Deepseekmath: Pushing the limits of mathematical reasoning in open language models},
  author={Shao, Zhihong and Wang, Peiyi and Zhu, Qihao and Xu, Runxin and Song, Junxiao and Bi, Xiao and Zhang, Haowei and Zhang, Mingchuan and Li, YK and Wu, Yang and others},
  journal={arXiv preprint arXiv:2402.03300},
  year={2024}
}

@article{schick2023toolformer,
  title={Toolformer: Language models can teach themselves to use tools},
  author={Schick, Timo and Dwivedi-Yu, Jane and Dess{\`\i}, Roberto and Raileanu, Roberta and Lomeli, Maria and Hambro, Eric and Zettlemoyer, Luke and Cancedda, Nicola and Scialom, Thomas},
  journal={Advances in neural information processing systems},
  volume={36},
  pages={68539--68551},
  year={2023}
}

@article{schulman2017proximal,
  title={Proximal policy optimization algorithms},
  author={Schulman, John and Wolski, Filip and Dhariwal, Prafulla and Radford, Alec and Klimov, Oleg},
  journal={arXiv preprint arXiv:1707.06347},
  year={2017}
}

@article{yu2026dapo,
  title={Dapo: An open-source llm reinforcement learning system at scale},
  author={Yu, Qiying and Zhang, Zheng and Zhu, Ruofei and Yuan, Yufeng and Zuo, Xiaochen and Yue, Yu and Dai, Weinan and Fan, Tiantian and Liu, Gaohong and Liu, Lingjun and others},
  journal={Advances in Neural Information Processing Systems},
  volume={38},
  pages={113222--113244},
  year={2026}
}

@article{zheng2025group,
  title={Group sequence policy optimization},
  author={Zheng, Chujie and Liu, Shixuan and Li, Mingze and Chen, Xiong-Hui and Yu, Bowen and Gao, Chang and Dang, Kai and Liu, Yuqiong and Men, Rui and Yang, An and others},
  journal={arXiv preprint arXiv:2507.18071},
  year={2025}
}

@article{rafailov2023direct,
  title={Direct preference optimization: Your language model is secretly a reward model},
  author={Rafailov, Rafael and Sharma, Archit and Mitchell, Eric and Manning, Christopher D and Ermon, Stefano and Finn, Chelsea},
  journal={Advances in neural information processing systems},
  volume={36},
  pages={53728--53741},
  year={2023}
}

@article{lambert2024tulu,
  title={Tulu 3: Pushing frontiers in open language model post-training},
  author={Lambert, Nathan and Morrison, Jacob and Pyatkin, Valentina and Huang, Shengyi and Ivison, Hamish and Brahman, Faeze and Miranda, Lester James V and Liu, Alisa and Dziri, Nouha and Lyu, Shane and others},
  journal={arXiv preprint arXiv:2411.15124},
  year={2024}
}

@article{liu2025understanding,
  title={Understanding r1-zero-like training: A critical perspective},
  author={Liu, Zichen and Chen, Changyu and Li, Wenjun and Qi, Penghui and Pang, Tianyu and Du, Chao and Lee, Wee Sun and Lin, Min},
  journal={arXiv preprint arXiv:2503.20783},
  year={2025}
}

@article{gao2025soft,
  title={Soft adaptive policy optimization},
  author={Gao, Chang and Zheng, Chujie and Chen, Xiong-Hui and Dang, Kai and Liu, Shixuan and Yu, Bowen and Yang, An and Bai, Shuai and Zhou, Jingren and Lin, Junyang},
  journal={arXiv preprint arXiv:2511.20347},
  year={2025}
}

@article{li2026skillsbench,
  title={SkillsBench: Benchmarking how well agent skills work across diverse tasks},
  author={Li, Xiangyi and Liu, Yimin and Chen, Wenbo and You, Bingran and Di, Zonglin and He, Yifeng and Zheng, Shenghan and Choe, Kyoung Whan and Sun, Jiankai and Wang, Shuyi and others},
  journal={arXiv preprint arXiv:2602.12670},
  year={2026}
}

@article{jiang2026sok,
  title={SoK: Agentic Skills--Beyond Tool Use in LLM Agents},
  author={Jiang, Yanna and Li, Delong and Deng, Haiyu and Ma, Baihe and Wang, Xu and Wang, Qin and Yu, Guangsheng},
  journal={arXiv preprint arXiv:2602.20867},
  year={2026}
}

@inproceedings{wang2026reinforcement,
  title={Reinforcement learning for self-improving agent with skill library},
  author={Wang, Jiongxiao and Yan, Qiaojing and Wang, Yawei and Tian, Yijun and Mishra, Soumya Smruti and Xu, Zhichao and Gandhi, Megha and Xu, Panpan and Cheong, Lin Lee},
  booktitle={Proceedings of the 64th Annual Meeting of the Association for Computational Linguistics (Volume 1: Long Papers)},
  pages={1529--1550},
  year={2026}
}

@article{xia2026skillrl,
  title={Skillrl: Evolving agents via recursive skill-augmented reinforcement learning},
  author={Xia, Peng and Chen, Jianwen and Wang, Hanyang and Liu, Jiaqi and Zeng, Kaide and Wang, Yu and Han, Siwei and Zhou, Yiyang and Zhao, Xujiang and Chen, Haifeng and others},
  journal={arXiv preprint arXiv:2602.08234},
  year={2026}
}

@article{yang2026skillopt,
  title={Skillopt: Executive strategy for self-evolving agent skills},
  author={Yang, Yifan and Gong, Ziyang and Huang, Weiquan and Yang, Qihao and Zhou, Ziwei and Huang, Zisu and Li, Yan and Gao, Xuemei and Dai, Qi and Liu, Bei and others},
  journal={arXiv preprint arXiv:2605.23904},
  year={2026}
}

@article{ni2026trace2skill,
  title={Trace2skill: Distill trajectory-local lessons into transferable agent skills},
  author={Ni, Jingwei and Liu, Yihao and Liu, Xinpeng and Sun, Yutao and Zhou, Mengyu and Cheng, Pengyu and Wang, Dexin and Zhao, Erchao and Jiang, Xiaoxi and Jiang, Guanjun},
  journal={arXiv preprint arXiv:2603.25158},
  year={2026}
}

@article{shen2026skillfoundry,
  title={Skillfoundry: Building self-evolving agent skill libraries from heterogeneous scientific resources},
  author={Shen, Shuaike and Cheng, Wenduo and Ma, Mingqian and Turcan, Alistair and Zhang, Martin Jinye and Ma, Jian},
  journal={arXiv preprint arXiv:2604.03964},
  year={2026}
}

@article{yang2026autoskill,
  title={Autoskill: Experience-driven lifelong learning via skill self-evolution},
  author={Yang, Yutao and Li, Junsong and Pan, Qianjun and Zhan, Bihao and Cai, Yuxuan and Du, Lin and Zhou, Jie and Chen, Kai and Chen, Qin and Li, Xin and others},
  journal={arXiv preprint arXiv:2603.01145},
  year={2026}
}

@article{wang2026skillx,
  title={Skillx: Automatically constructing skill knowledge bases for agents},
  author={Wang, Chenxi and Yu, Zhuoyun and Xie, Xin and Yao, Wuguannan and Fang, Runnan and Qiao, Shuofei and Cao, Kexin and Zheng, Guozhou and Qi, Xiang and Zhang, Peng and others},
  journal={arXiv preprint arXiv:2604.04804},
  year={2026}
}

@inproceedings{fang2026memp,
  title={Memp: Exploring agent procedural memory},
  author={Fang, Runnan and Liang, Yuan and Wang, Xiaobin and Wu, Jialong and Qiao, Shuofei and Xie, Pengjun and Huang, Fei and Chen, Huajun and Zhang, Ningyu},
  booktitle={Findings of the Association for Computational Linguistics: ACL 2026},
  pages={17490--17502},
  year={2026}
}

@article{alzubi2026evoskill,
  title={Evoskill: Automated skill discovery for multi-agent systems},
  author={Alzubi, Salaheddin and Provenzano, Noah and Bingham, Jaydon and Chen, Weiyuan and Vu, Tu},
  journal={arXiv preprint arXiv:2603.02766},
  year={2026}
}

@article{liu2026skillforge,
  title={Skillforge: Forging domain-specific, self-evolving agent skills in cloud technical support},
  author={Liu, Xingyan and Luo, Xiyue and Li, Linyu and Huang, Ganghong and Liu, Jianfeng and Qiao, Honglin},
  journal={arXiv preprint arXiv:2604.08618},
  year={2026}
}

@article{zhang2026coevoskills,
  title={Coevoskills: Self-evolving agent skills via co-evolutionary verification},
  author={Zhang, Hanrong and Fan, Shicheng and Zou, Henry Peng and Chen, Yankai and Wang, Zhenting and Zhou, Jiayu and Li, Chengze and Huang, Wei-Chieh and Yao, Yifei and Zheng, Kening and others},
  journal={arXiv preprint arXiv:2604.01687},
  year={2026}
}

@article{ma2026skillclaw,
  title={Skillclaw: Let skills evolve collectively with agentic evolver},
  author={Ma, Ziyu and Yang, Shidong and Ji, Yuxiang and Wang, Xucong and Wang, Yong and Hu, Yiming and Huang, Tongwen and Chu, Xiangxiang},
  journal={arXiv preprint arXiv:2604.08377},
  year={2026}
}

@article{qiu2026autorefine,
  title={AutoRefine: From Trajectories to Reusable Expertise for Continual LLM Agent Refinement},
  author={Qiu, Libin and Gao, Zhirong and Chen, Junfu and Ye, Yuhang and Huang, Weizhi and Xue, Xiaobo and Qiu, Wenkai and Tang, Shuo},
  journal={arXiv preprint arXiv:2601.22758},
  year={2026}
}

@article{mi2026procmem,
  title={ProcMEM: Learning Reusable Procedural Memory from Experience via Non-Parametric PPO for LLM Agents},
  author={Mi, Qirui and Ma, Zhijian and Yang, Mengyue and Li, Haoxuan and Wang, Yisen and Zhang, Haifeng and Wang, Jun},
  journal={arXiv preprint arXiv:2602.01869},
  year={2026}
}

@article{wu2025evolver,
  title={Evolver: Self-evolving llm agents through an experience-driven lifecycle},
  author={Wu, Rong and Wang, Xiaoman and Mei, Jianbiao and Cai, Pinlong and Fu, Daocheng and Yang, Cheng and Wen, Licheng and Yang, Xuemeng and Shen, Yufan and Wang, Yuxin and others},
  journal={arXiv preprint arXiv:2510.16079},
  year={2025}
}

@article{bischl2023hyperparameter,
  title={Hyperparameter optimization: Foundations, algorithms, best practices, and open challenges},
  author={Bischl, Bernd and Binder, Martin and Lang, Michel and Pielok, Tobias and Richter, Jakob and Coors, Stefan and Thomas, Janek and Ullmann, Theresa and Becker, Marc and Boulesteix, Anne-Laure and others},
  journal={Wiley Interdisciplinary Reviews: Data Mining and Knowledge Discovery},
  volume={13},
  number={2},
  pages={e1484},
  year={2023},
  publisher={Wiley Online Library}
}

@article{yang2022click,
  title={Click-through rate prediction in online advertising: A literature review},
  author={Yang, Yanwu and Zhai, Panyu},
  journal={Information Processing \& Management},
  volume={59},
  number={2},
  pages={102853},
  year={2022},
  publisher={Elsevier}
}

@inproceedings{chapelle2014modeling,
  title={Modeling delayed feedback in display advertising},
  author={Chapelle, Olivier},
  booktitle={Proceedings of the 20th ACM SIGKDD international conference on Knowledge discovery and data mining},
  pages={1097--1105},
  year={2014}
}

@inproceedings{mcdonald2023impatient,
  title={Impatient bandits: Optimizing recommendations for the long-term without delay},
  author={McDonald, Thomas M and Maystre, Lucas and Lalmas, Mounia and Russo, Daniel and Ciosek, Kamil},
  booktitle={Proceedings of the 29th ACM SIGKDD Conference on Knowledge Discovery and Data Mining},
  pages={1687--1697},
  year={2023}
}

@inproceedings{yu2024addressing,
  title={Addressing delayed feedback in conversion rate prediction: A domain adaptation approach},
  author={Yu, Leisheng and Cai, Yanxiao and Chen, Lucas and Zhang, Minxing and Day, Wei-Yen and Li, Li and Chen, Rui and Choi, Soo-Hyun and Hu, Xia},
  booktitle={2024 IEEE International Conference on Data Mining (ICDM)},
  pages={917--922},
  year={2024},
  organization={IEEE}
}

@inproceedings{zhang2021counterfactual,
  title={Counterfactual reward modification for streaming recommendation with delayed feedback},
  author={Zhang, Xiao and Jia, Haonan and Su, Hanjing and Wang, Wenhan and Xu, Jun and Wen, Ji-Rong},
  booktitle={Proceedings of the 44th international ACM SIGIR conference on research and development in information retrieval},
  pages={41--50},
  year={2021}
}

@inproceedings{yasui2020feedback,
  title={A feedback shift correction in predicting conversion rates under delayed feedback},
  author={Yasui, Shota and Morishita, Gota and Komei, Fujita and Shibata, Masashi},
  booktitle={Proceedings of the Web Conference 2020},
  pages={2740--2746},
  year={2020}
}

@inproceedings{guo2023leveraging,
  title={Leveraging post-click user behaviors for calibrated conversion rate prediction under delayed feedback in online advertising},
  author={Guo, Yuyao and Ao, Xiang and Liu, Qiming and He, Qing},
  booktitle={Proceedings of the 32nd ACM International Conference on Information and Knowledge Management},
  pages={3918--3922},
  year={2023}
}

@article{yang2022generalized,
  title={Generalized delayed feedback model with post-click information in recommender systems},
  author={Yang, Jiaqi and Zhan, De-Chuan},
  journal={Advances in Neural Information Processing Systems},
  volume={35},
  pages={26192--26203},
  year={2022}
}

@inproceedings{su2021attention,
  title={An attention-based model for conversion rate prediction with delayed feedback via post-click calibration},
  author={Su, Yumin and Zhang, Liang and Dai, Quanyu and Zhang, Bo and Yan, Jinyao and Wang, Dan and Bao, Yongjun and Xu, Sulong and He, Yang and Yan, Weipeng},
  booktitle={Proceedings of the Twenty-Ninth International Conference on International Joint Conferences on Artificial Intelligence},
  pages={3522--3528},
  year={2021}
}

@inproceedings{gu2021real,
  title={Real negatives matter: Continuous training with real negatives for delayed feedback modeling},
  author={Gu, Siyu and Sheng, Xiang-Rong and Fan, Ying and Zhou, Guorui and Zhu, Xiaoqiang},
  booktitle={Proceedings of the 27th ACM SIGKDD Conference on Knowledge Discovery \& Data Mining},
  pages={2890--2898},
  year={2021}
}

@inproceedings{chen2022asymptotically,
  title={Asymptotically unbiased estimation for delayed feedback modeling via label correction},
  author={Chen, Yu and Jin, Jiaqi and Zhao, Hui and Wang, Pengjie and Liu, Guojun and Xu, Jian and Zheng, Bo},
  booktitle={Proceedings of the ACM Web Conference 2022},
  pages={369--379},
  year={2022}
}

@inproceedings{wang2023unbiased,
  title={Unbiased delayed feedback label correction for conversion rate prediction},
  author={Wang, Yifan and Sun, Peijie and Zhang, Min and Jia, Qinglin and Li, Jingjie and Ma, Shaoping},
  booktitle={Proceedings of the 29th ACM SIGKDD Conference on Knowledge Discovery and Data Mining},
  pages={2456--2466},
  year={2023}
}

@inproceedings{yasui2022learning,
  title={Learning classifiers under delayed feedback with a time window assumption},
  author={Yasui, Shota and Kato, Masahiro},
  booktitle={Proceedings of the 28th ACM SIGKDD Conference on Knowledge Discovery and Data Mining},
  pages={2286--2295},
  year={2022}
}

@inproceedings{yang2021capturing,
  title={Capturing delayed feedback in conversion rate prediction via elapsed-time sampling},
  author={Yang, Jia-Qi and Li, Xiang and Han, Shuguang and Zhuang, Tao and Zhan, De-Chuan and Zeng, Xiaoyi and Tong, Bin},
  booktitle={Proceedings of the AAAI Conference on Artificial Intelligence},
  volume={35},
  number={5},
  pages={4582--4589},
  year={2021}
}

@inproceedings{hou2021conversion,
  title={Conversion prediction with delayed feedback: A multi-task learning approach},
  author={Hou, Yilin and Zhao, Guangming and Liu, Chuanren and Zu, Zhonglin and Zhu, Xiaoqiang},
  booktitle={2021 IEEE International Conference on Data Mining (ICDM)},
  pages={191--199},
  year={2021},
  organization={IEEE}
}

@article{wu2026agenticrectune,
  title={AgenticRecTune: Multi-Agent with Self-Evolving Skillhub for Recommendation System Optimization},
  author={Wu, Xidong and Zhuan, Yue and Wei, Ruoqiao and Chen, Hangxin and Bai, Di and Liu, Jintao and Wang, Xinyi and Wang, Xue and Wang, Luoshu and Cheng, Xinwu},
  journal={arXiv preprint arXiv:2604.26969},
  year={2026}
}

@inproceedings{deng2013improving,
  title={Improving the sensitivity of online controlled experiments by utilizing pre-experiment data},
  author={Deng, Alex and Xu, Ya and Kohavi, Ron and Walker, Toby},
  booktitle={Proceedings of the sixth ACM international conference on Web search and data mining},
  pages={123--132},
  year={2013}
}

@inproceedings{ganin2015unsupervised,
  title={Unsupervised domain adaptation by backpropagation},
  author={Ganin, Yaroslav and Lempitsky, Victor},
  booktitle={International conference on machine learning},
  pages={1180--1189},
  year={2015},
  organization={PMLR}
}

@techreport{athey2019surrogate,
  title={The surrogate index: Combining short-term proxies to estimate long-term treatment effects more rapidly and precisely},
  author={Athey, Susan and Chetty, Raj and Imbens, Guido W and Kang, Hyunseung},
  year={2019},
  institution={National Bureau of Economic Research}
}

@article{lim2021temporal,
  title={Temporal fusion transformers for interpretable multi-horizon time series forecasting},
  author={Lim, Bryan and Ar{\i}k, Sercan {\"O} and Loeff, Nicolas and Pfister, Tomas},
  journal={International journal of forecasting},
  volume={37},
  number={4},
  pages={1748--1764},
  year={2021},
  publisher={Elsevier}
}

\clearpage
\beginappendix

\subsection{Ethical Considerations}

Our work on DMRL aims to advance automated skill optimization in advertising recommendation under noisy, delayed, and heterogeneous feedback. While this capability can improve the efficiency and robustness of industrial recommendation systems, we recognize that, as with automated decision-making framework operating on user data, it is essential to consider the potential ethical implications, including fairness, privacy, and broader societal impact.

Fairness and Bias. DMRL explicitly models population heterogeneity through LRP, which improves long-term value estimation across diverse user groups, but it also raises the risk that automated optimization systematically favor segments with higher monetization potential or more predictable response patterns. In deploying DMRL, we mitigate such risks through segment-wise evaluation and constraint-aware monitoring throughout both offline analysis and online experimentation. When significant imbalances are detected, optimization process is restricted before broader deployment.

User Privacy. The LRP module is trained offline using platform data, making privacy-preserving data handling a necessary requirement of the system. In practice, DMRL relies on anonymized logs, controlled feature access, and strict internal data-retention policies in accordance with platform data-governance standards. Sensitive identifiers are protected through standard anonymization procedures, and long-term reward modeling is based on aggregated behavioral signals. In addition, training data access and model artifacts are subject to internal review and management controls to reduce the risk of exposing personally identifiable information.

Societal Impact. DMRL enhances the ability of advertising systems to optimize decisions automatically and at scale, especially through DRPO, which operates with online training and feedback-driven optimization. While such automation can substantially improve system efficiency, it may also increase the risk that optimization becomes overly focused on monetization, potentially affecting user autonomy, advertiser balance, or content diversity. To address this concern, DMRL is deployed with practical operational safeguards, including bounded policy rollout, online anomaly monitoring, and business-rule constraints on optimization updates.

Mitigation and Future Work. This research is conducted with these ethical considerations embedded into both system training and deployment. DMRL already incorporates segment-wise monitoring, privacy-compliant data handling, and controlled rollout procedures to mitigate bias, privacy risk, and harmful optimization. Looking ahead, we strengthen these safeguards by introducing fairness-aware objectives, privacy-enhancing techniques, and more transparent auditing tools for long-term optimization effects.

\end{document}